\pdfoutput=1

\documentclass[11pt]{article}

\usepackage[]{acl}

\usepackage{times}
\usepackage{latexsym}
\usepackage{multicol}
\usepackage{multirow}
\usepackage{graphicx}
\usepackage[T1]{fontenc}

\usepackage{amsmath}

\usepackage{tikz}
\newcommand*\circled[1]{\tikz[baseline=(char.base)]{
            \node[shape=circle,draw,inner sep=0.8pt] (char) {#1};}}

\usepackage[utf8]{inputenc}

\usepackage{microtype}

\usepackage{inconsolata}

\title{Hierarchical Organization Simulacra in the Investment Sector}

\author{Chung-Chi Chen,\textsuperscript{1} Hiroya Takamura,\textsuperscript{1} Ichiro Kobayashi,\textsuperscript{2}  Yusuke Miyao\textsuperscript{3}
\\
 \textsuperscript{1} Artificial Intelligence Research Center, AIST, Japan \\
\textsuperscript{2} Ochanomizu University, Japan \\
 \textsuperscript{3} University of Tokyo, Japan \\
   \texttt{c.c.chen@acm.org, takamura.hiroya@aist.go.jp,}\\ \texttt{koba@is.ocha.ac.jp, yusuke@is.s.u-tokyo.ac.jp}\\
}
\begin{document}
\maketitle
\begin{abstract}
This paper explores designing artificial organizations with professional behavior in investments using a multi-agent simulation. The method mimics hierarchical decision-making in investment firms, using news articles to inform decisions. A large-scale study analyzing over 115,000 news articles of 300 companies across 15 years compared this approach against professional traders' decisions. Results show that hierarchical simulations align closely with professional choices, both in frequency and profitability. However, the study also reveals biases in decision-making, where changes in prompt wording and perceived agent seniority significantly influence outcomes. This highlights both the potential and limitations of large language models in replicating professional financial decision-making.
\end{abstract}

\section{Introduction}

How might we design an artificial organization that exemplifies professional behavior? With the advent of large language models (LLMs), increasing discussions have emerged on this topic, such as software development~\cite{qian2023communicative} and text evaluation~\cite{chan2023chateval}. These studies have primarily focused on enabling models to discuss various aspects, yet they overlook the bottom-up decision-making process. In some business environments, organizations are hierarchical, with final decisions made by managers or executives. Following this line of thought, this paper aims to bridge this gap by exploring hierarchical organizations in the investment sector.

\begin{figure}[t]
  \centering
  \includegraphics[width=5.5cm]{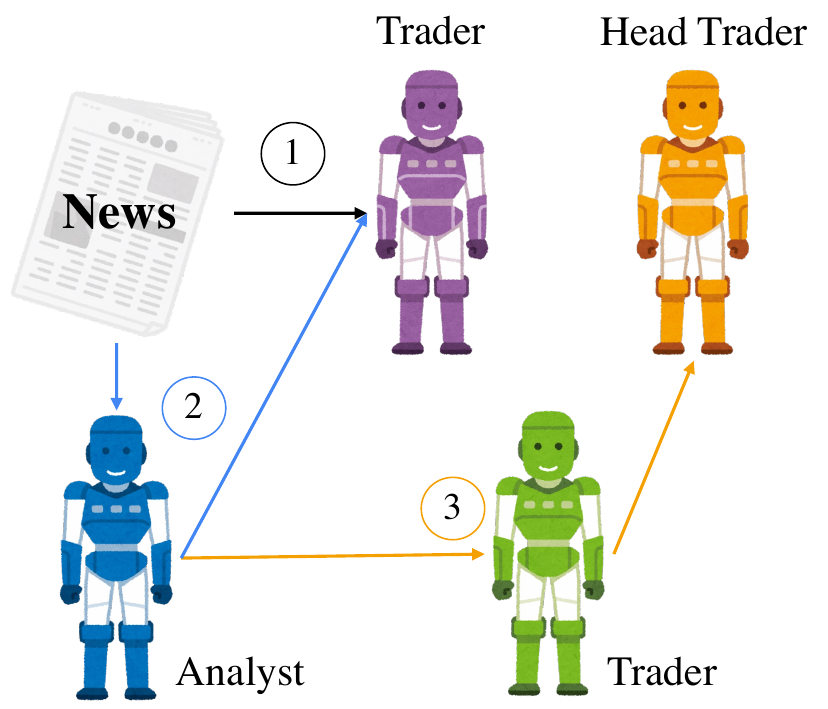}
  \caption{Illustration of communication strategies. }
  \label{fig:Example}
\end{figure}

Figure~\ref{fig:Example} illustrates various designs of communication strategies. For instance, strategy (1) involves a single agent directly making a decision. Strategy (2) entails disseminating information to two agents for analysis from distinct perspectives before finalizing the decision. Strategy (3), emulating hierarchical organizations, introduces an additional agent responsible for approving decisions made by the agents in strategy \circled{2}.
In this paper, we propose and evaluate a multi-agent simulation approach that mirrors the hierarchical decision-making processes typically found in investment firms. Aligning with the findings of previous studies that highlight the significance of news in financial forecasting~\cite{peng-jiang-2016-leverage, chen-etal-2019-incorporating, li2021modeling, li2023pen}, we introduce a conceptual framework wherein news articles are used as a trigger for investment decisions. 
Our experiment involves a multi-layered approach where different agents, powered by LLMs, interact in a sequence that mimics the layers of decision-making in a traditional trading environment. 
Our study revolves around assessing the effectiveness of this hierarchical multi-agent setup in replicating the decision-making of professionals in the field of investment.

The primary research question in this paper concerns the efficacy of a multi-agent simulation approach in mirroring the decision-making processes of professionals in the field. To this end, we collected trading records from financial institutions on the stock exchange, enabling a comparison between the decisions of these professionals and LLMs. To bolster the robustness of our discussion, we undertook a large-scale experiment, analyzing over 115K news articles pertaining to 300 companies over a 15-year span. Our results indicate that implementing a hierarchical multi-agent approach not only increases the frequency of decisions that are consistent with those made by traders in financial institutions but also yields more profitable outcomes compared to the conventional multi-agent approach. Our findings also reveal that modifying a single concept in the prompt can significantly change the decisions. For instance, altering short-term to long-term in the prompt resulted in decisions more closely aligned with professional traders. Another notable observation is the preference of the head trader for decisions from the ``senior'' trader over those from the ``junior'' trader, despite identical rationale in the prompt. This highlights an inherent bias in the multi-agent approach, where the LLM favors decisions based on perceived seniority.

\section{Related Work}
The application of a multi-agent approach across diverse research domains has recently gained traction. \citet{yao2022webshop} employed this approach to simulate web browsing behavior in humans. \citet{park2023generative} extended its application to mimic human behaviors in social intersections, while \citet{zhang2023exploring} focused on the psychological aspects. Beyond simulating human behavior, this approach has been instrumental in addressing specific tasks. \citet{qian2023communicative} explored communicative dynamics in software development, highlighting how LLMs enhance team communication. \citet{chan2023chateval} concentrated on text evaluation, specifically assessing content quality and relevance. These investigations form a foundation for understanding the role of multi-agent approaches in solving tasks. 
Our study advances this body of work by applying LLMs within a hierarchical decision-making framework, incorporating external news sources for financial forecasting, and examining the dynamics in simulated professional investment environments. This multidisciplinary strategy offers a broader perspective on the effective employment of LLMs in hierarchical organizational structures.

\section{Dataset}
In our experiment, we utilized two datasets: news articles from news vendors and trading records from financial institutions obtained from the stock exchange.\footnote{For the details on dataset reconstruction and data vendors, please refer to Appendix~\ref{rebuild dataset}.}
Three prominent news sources in Taiwan were selected for this study: the Market Observation Post System, the Economic Daily News, and the Commercial Times. We acquired 115,223 news articles pertaining to 300 companies. These articles span a period of 15 years, from July 2004 to March 2020.

The Taiwan Stock Exchange publicly discloses the daily trading records of financial institutions. These records include the net buying or selling amounts by professional institutions for each stock. We correlated these trading records with the news articles, generating labels that indicate whether professional institutions' actions on a given stock were overweight, underweight, or neutral. Here, overweight (underweight) implies that the total buying (selling) volume exceeded the selling (buying) volume, and neutral denotes that no professional institution trades on the given stock. During the 15-year period, the proportions of these three categories were 39.19\%, 39.29\%, and 21.52\%, respectively.

\section{Interactive Simulacra}
This paper aims to determine the extent to which agents' decisions (overweight, underweight, and neutral) align with those of professional institutions. This necessitates design approaches that categorize agents into one of these three decision-making types for evaluation.

Our methodology involves using news as a stimulus. 
Our framework encompasses three roles: 
\textbf{Analyst}: Delivers analyses of potential future outcomes, both positive and negative, derived from the provided news.
\textbf{Trader}: Decides based on the available information (news or the analyst's analysis) and articulates the rationales.
\textbf{Head Trader}: Determines whether to follow the trader's recommendations to overweight or underweight. 

The conventional method entails feeding news as input, subsequently prompting the trader agent to decide whether to buy or sell a company's stock on the day following the news release. This approach is referred to as the Single Trader strategy in our study.
Inspired by the chain-of-thought (CoT)~\cite{wei2022chain}, we hypothesize that including analysis generated by an agent may enhance performance. To test this, we introduce an analyst agent into the process. This allows the trader agent to base its decision on the analyst's analysis. Furthermore, we incorporate a head trader agent to mirror the hierarchical structure commonly observed in trading desks, where the head trader typically makes the final decisions. This method is denoted as the Hierarchical Organization (HO) strategy in our study.
To expand our research, we introduce an additional trader agent, supported by an LLM different from the first. This is to investigate whether augmenting various input leads to better outcomes. We designate this variation as $\text{HO}^m$.
In the multiple-trader scenario, the head trader decides whether to follow one of the traders' recommendations or neither.

This paper primarily employs GPT-3.5 to perform various roles. In the two-trader scenario within the hierarchical organization approach, we use PaLM-2\footnote{\url{https://ai.google/discover/palm2/}} as the other trader.
The prompts of different agents are provided in the Appendix~\ref{Prompts}.

\begin{table}[t]
  \centering
  \small
    \begin{tabular}{l|rrr}
          & \multicolumn{1}{c}{Overweight} & \multicolumn{1}{c}{Neutral} & \multicolumn{1}{c}{Underweight}  \\
    \hline
    Institutions & 39.19\% & 21.52\% & 39.29\% \\
    \hline
    Single Trader & 19.86\% & 70.27\% & 9.87\% \\
    CoT & 17.87\% & 72.02\% & 10.12\% \\
    HO & 11.81\% & 79.14\% & 9.05\% 
    \end{tabular}%
  \caption{Statistics of decision.}
  \label{tab:Statistics of decisions.}%
\end{table}%

\begin{table}[t]
  \centering
    \small
 \begin{tabular}{l|r|r|r}
          & \multicolumn{1}{c|}{Overall} & \multicolumn{1}{c|}{Overweight} & \multicolumn{1}{c}{Underweight} \\
    \hline
    Single Trader & 42.64\% & 44.41\% & \textbf{39.08\%} \\
    CoT   & 42.88\% & 43.95\% & 34.72\% \\
    HO    & \textbf{44.77\%} & \textbf{45.09\%} & 32.51\% \\
    \end{tabular}%
  \caption{Consistency between agents and financial institutions.}
  \label{tab:Consistency}%
\end{table}%

\begin{table}[t]
  \centering
  \small
    \begin{tabular}{l|r|r|r}
          & \multicolumn{1}{c|}{Overall} & \multicolumn{1}{c|}{Overweight} & \multicolumn{1}{c}{Underweight} \\
    \hline
    Short-Term & 38.51\% & \textbf{44.03\%} & 28.75\% \\
    Long-Term & \textbf{42.88\%} & 43.95\% & \textbf{34.72\%} \\
    \end{tabular}%
  \caption{Consistency of CoT with different temporal settings. }
  \label{tab:temporal}%
\end{table}%

\begin{table}[t]
  \centering
  \resizebox{\columnwidth}{!}{
    \begin{tabular}{rc|rrr|r}
          &       & \multicolumn{3}{c|}{Long-Term} &  \\
          &       & \multicolumn{1}{c}{Overweight} & \multicolumn{1}{c}{Neutral} & \multicolumn{1}{c|}{Underweight} & \multicolumn{1}{c}{Overall} \\
    \hline
     \multirow{3}[0]{*}{\rotatebox{90}{Short-T.}} & Overweight  & 15.15\% & 3.05\% & 0.01\% & 18.21\% \\
          & Neutral & 6.18\% & 65.72\% & 0.38\% & 72.29\% \\
          & Underweight & 0.06\% & 7.10\% & 2.34\% & 9.51\% \\
    \hline
          & Total & 21.39\% & 75.87\% & 2.73\% &  \\
    \end{tabular}%
    }
  \caption{Comparison of short-/long-term decisions.}
  \label{tab:Comparison of short-/long-term decisions}%
\end{table}%

\section{Experiment and Discussion}
\subsection{Statistics of Decision}
In Table~\ref{tab:Statistics of decisions.}, we present statistics on the decisions of financial institutions and agents. Our findings indicate that agents are more conservative, making overweight or underweight decisions in approximately 30\% of instances, whereas financial institutions take action in about 80\% of cases. Consequently, our subsequent discussions focus on the ability of agents to replicate the decisions (either overweight or underweight) made by professionals.
In the following section, we use \textit{consistency}, the ratio of the number of identical decisions made by both agents and institutions to the number of agents' overweight/underweight decisions, for comparisons. 
This is mathematically expressed as
$\it{Consistency} = \frac{\text{\# identical decisions made by both agents and institutions}}{\text{\# agents' overweight/underweight decisions}}$

\subsection{Consistency with Financial Institutions}
Table~\ref{tab:Consistency} shows that the HO strategy is superior in aligning with professional decisions, while the CoT strategy has a lower probability of mirroring professional decisions. 
These results indicate that replicating the organizational dynamics of professional entities in decision-making simulations can improve the congruence between agents and experts. 
Further analysis reveals that, among the three strategies, the HO strategy achieves the highest consistency with professionals in making overweight decisions, both in terms of matching decisions and avoiding contradictory ones. Conversely, for underweight decisions, the frequency of aligning with professional decisions is lower across all strategies compared to overweight decisions.

\subsection{Temporal-Aware Decision}
In this section, we examine the impact of varying temporal durations (short-term and long-term) on investment consistency. We modified the prompt given to the trading agent, instructing it to make decisions over periods of one week and one year. As depicted in Table~\ref{tab:temporal}, longer-term considerations enhance consistency, particularly for underweight recommendations. 
Additionally, we juxtapose short-term and long-term decisions in Table~\ref{tab:Comparison of short-/long-term decisions}. 
We find that trading agents with a long-term perspective tend to favor overweight decisions and reduce underweight ones. This shift improves the consistency of underweight decisions while marginally diminishing it in overweight decisions. These results suggest that the decisions of LLMs over extended periods align more closely with those of professionals.

\subsection{Seniority-Aware Decision}
In this section, we explore the head trader's decisions by altering a single word in the prompt. Specifically, with identical decisions and rationales, we inform the head trader agent that the advice originates from either a junior or a senior trader. Table~\ref{tab:Head trader bias toward seniority.} illustrates that the head trader agent is more inclined to accept decisions presented as emanating from a senior trader. We also demonstrate that this seniority bias undermines overall consistency, particularly in decisions of overweighting. Conversely, a lower acceptance rate for decisions described as coming from junior traders contributes to improved consistency in underweight decisions, as evidenced by the low consistency of trader agents in such decisions in Table~\ref{tab:Consistency}. Our results indicate that a bias based on seniority is apt to emerge at the execution level, and it subsequently impacts the decisions.

\begin{table}[t]
  \centering
  \resizebox{\columnwidth}{!}{
    \begin{tabular}{l|r|rrr}
          & \multicolumn{1}{c|}{Approval Rate} & \multicolumn{1}{c}{Overall} & \multicolumn{1}{c}{Overweight} & \multicolumn{1}{c}{Underweight} \\
    \hline
    Junior & 21.56\% & 44.16\% & 44.28\% & \textbf{37.90\%} \\
    Senior & 38.65\% & \textbf{44.77\%} & \textbf{45.09\%} & 32.51\% \\
    \end{tabular}%
    }
  \caption{Head trader bias toward seniority and its influence on consistency.}
  \label{tab:Head trader bias toward seniority.}%
\end{table}%

\begin{table}[t]
  \centering
  \resizebox{\columnwidth}{!}{
    \begin{tabular}{l|rrr}
          & \multicolumn{1}{c}{Overall} & \multicolumn{1}{c}{Overweight} & \multicolumn{1}{c}{Underweight} \\
    \hline
    CoT (GPT-3.5) & \textbf{42.88\%} & 43.95\% & 34.72\% \\
    CoT (PaLM-2) & 41.55\% & 42.03\% & \textbf{41.05\%} \\
    $\text{HO}^m \text{(GPT-3.5)}$ & 41.74\% & 43.79\% & 37.28\% \\
    $\text{HO}^m \text{(PaLM-2)}$ & 42.00\% & \textbf{44.36\%} & 37.49\% \\
    \end{tabular}%
    }
  \caption{Consistency of multiple trader strategy.}
  \label{tab:palm}%
\end{table}%

\subsection{Considering Multiple Opinions}
This section evaluates the impact of incorporating an additional trader with a different LLM on performance outcomes. Table~\ref{tab:palm} initially contrasts the efficacy of PaLM-2 as a trader, specifically CoT (PaLM-2), against GPT-3.5. The findings indicate superior overall performance with GPT-3.5, while PaLM-2 aligns closely with financial institutions in underweight. Further, we introduce both GPT-3.5 and PaLM-2 as head traders in the $\text{HO}^m$ strategy, yielding dual perspectives: $\text{HO}^m \text{(GPT-3.5)}$ and $\text{HO}^m \text{(PaLM-2)}$. The results suggest enhanced overall consistency when PaLM-2 leads, but a reduction when GPT-3.5 serves as the head trader. Notably, the consistency of $\text{HO}^m \text{(GPT-3.5)}$ in underweight decisions surpasses that of HO, which excludes the viewpoint of the PaLM-2 trader. This implies that leveraging diverse LLMs in interaction simulations may facilitate a multifaceted understanding, potentially enhancing alignment with professional investment standards.

\begin{table}[t]
  \centering
  \small
    \begin{tabular}{llrr}
    \multicolumn{1}{c}{Decision} & \multicolumn{1}{c}{Strategy} & \multicolumn{1}{c}{$t+1$} & \multicolumn{1}{c}{$t+5$} \\
    \hline
    \multirow{6}[4]{*}{Overweight} & Single Trader & 57.10\% & 51.96\% \\
          & CoT (Short-Term) & 58.51\% & 52.49\% \\
          & CoT (Long-Term) & 56.37\% & 51.99\% \\
          & HO (Junior) & \textbf{60.55\%} & 51.77\% \\
          & HO (Senior) & 59.93\% & \textbf{53.49\%} \\
          & $\text{HO}^m \text{(GPT-3.5)}$ & 58.50\% & 53.24\% \\ 
          & $\text{HO}^m \text{(PaLM-2)}$ & 58.50\% & 52.28\% \\ 
\cline{2-4}          & Institutions & 55.30\% & 57.08\% \\
    \hline
    \multirow{6}[3]{*}{Underweight} & Single Trader & 40.81\% & 53.16\% \\
          & CoT (Short-Term) & 40.23\% & 53.25\% \\
          & CoT (Long-Term) & \textbf{41.27\%} & 51.09\% \\
          & HO (Junior) & 37.10\% & 46.77\% \\
          & HO (Senior) & 37.86\% & 52.26\% \\
          & $\text{HO}^m \text{(GPT-3.5)}$ & 40.54\% & \textbf{53.38\%} \\
          & $\text{HO}^m \text{(PaLM-2)}$ & 40.73\% & 53.24\% \\ 
\cline{2-4}          & Institutions & 32.04\% & 54.38\% \\
    \end{tabular}%
  \caption{Consistency between agents and market.}
  \label{tab:Consistency market}%
\end{table}%

\section{Consistency with the Market}
In addition to alignment with professional financial institutions, this study also evaluates the congruence between market price movements at day $t+T$ and decisions made at day $t$. Specifically, if a decision to overweight coincides with a subsequent rise in market price at day $t+T$, it is deemed an aligned decision. It is important to clarify that our objective is to understand the extent of alignment between decisions and market movements.

Table~\ref{tab:Consistency market} indicates that the HO strategy exhibits the highest consistency with market movements, with 60.55\% of overweight decisions followed by an upward price movement. Notably, the consistency at $t+1$ surpasses that at $t+5$ across all strategies. This trend contrasts with that observed in professional institutions, where consistency is greater at $t+5$. Regarding underweight decisions, all strategies, including those of institutions, do not exhibit good consistency with market movements at $t+1$.
However, over a longer period ($t+5$), the $\text{HO}^m$ strategy with GPT-3.5 as the head trader achieves the highest consistency among all strategies. 
Professional institutions also demonstrate higher consistency over this extended duration.

\section{Conclusion}
This paper has delved into the potential of hierarchical interactive simulacra in the investment sector, particularly exploring how LLMs can simulate professional investment decision-making processes. 
The insights gained from the hierarchical multi-agent approach, the impact of seniority biases, and the integration of multiple LLMs offer several perspectives for future research and practical applications in the field of finance and beyond. The promising alignment with market movements further emphasizes the potential of LLMs in enhancing decision-making processes in professional settings.

\section*{Limitations}
First, the exclusive reliance on news articles for investment decisions fails to encapsulate the multifaceted factors influencing real-world trading. Professional traders typically evaluate a wider array of information, including market trends, economic indicators, and specific company data, aspects not entirely captured in our model.
Second, while the hierarchical decision-making framework presents innovation, it may oversimplify the dynamics within actual investment firms. Investment decisions in reality often result from collaborative discussions among multiple stakeholders with varied expertise, a situation our hierarchical model does not entirely replicate.
Additionally, the concentration of our experiment on the Taiwan market constrains the applicability of our results. Financial markets exhibit considerable regional variation, and the behaviors observed in this study may not reflect trends in other global financial contexts.
Finally, the study's emphasis on alignment with market movements and professional decisions might neglect other vital trading aspects, such as risk management, ethical considerations, and regulatory compliance. These elements are essential for responsible investment practices.
Future research should strive to mitigate these limitations by integrating more diverse data sources, examining more intricate decision-making frameworks, extending the research to various markets, and including additional trading facets beyond mere alignment with market dynamics.

\section*{Impact Statement}
In this paper, we examine the role of models in guiding investment behavior. While these models show promise in enhancing decision-making, it is vital to recognize the inherent risks associated with this line of research. Understanding and managing these risks is imperative, especially when implementing AI systems in financially sensitive sectors. Despite the potential benefits, the risks must be concurrently evaluated and addressed.
Our objective is to deepen the understanding of LLMs' decision-making processes, rather than to forecast future events. Readers should be aware that even minor variations in prompts can lead to substantially different outcomes. Therefore, the recommendations provided by the methods discussed herein should not be applied directly to real-world investment scenarios. Investment is a specialized field, and it is advisable to consult licensed analysts before making decisions and undertaking any actions.

\bibliography{custom}

\appendix

\section{Rebuilding Dataset}
\label{rebuild dataset}
Three prominent news sources in Taiwan were selected for this study: the Market Observation Post System,\footnote{\url{https://emops.twse.com.tw/}} the Economic Daily News,\footnote{\url{https://money.udn.com/money/}} and the Commercial Times.\footnote{\url{https://www.ctee.com.tw/}}
The daily trading record can be obtained from the Taiwan Stock Exchange's page.\footnote{\url{https://www.twse.com.tw/en/trading/foreign/t86.html}}

We provide news headlines, labels, and agent-generated text for future research. Due to ethical concerns, we do not include the full news content. However, researchers can reconstruct and access the complete dataset using the provided details through FinMind,\footnote{\url{https://finmind.github.io/}} an open-source dataset. The comprehensive news content is retrievable by querying the API with the date, news title, and stock ticker information we have supplied.

\section{Prompts}
\label{Prompts}
Table~\ref{tab:prompts} provides the prompts we used in the experiments. 
In the temporal concept and seniority changing experiments, we only change the \textcolor{red}{red} text: (1) use ``Our strategy is to hold the position for a year, considering long-term effects and potential price movements.'' replaces ``Our objective is to capitalize on potential market movements within the upcoming week. To aid our decision, it's vital to gauge whether the recent news will significantly influence investor views and cause short-term price fluctuations.'', and (2) use ``senior'' replaces ``junior.''

\begin{table*}[t]
  \centering
  \resizebox{\textwidth}{!}{
    \begin{tabular}{l|l|p{35em}}
    \multicolumn{1}{c|}{Role} & \multicolumn{1}{c|}{Input Information} & \multicolumn{1}{c}{Prompt} \\
    \hline
    \multirow{8}[4]{*}{Analyst} & \multirow{8}[4]{*}{News}  & Based on the following news, please give me some potential positive/negative scenarios for the mentioned companies. \newline{}\newline{}News Title: \newline{}{\{News Title\}}\newline{}\newline{}News Content: \newline{}{\{News Content\}} \\
    \hline
    \multirow{14}[7]{*}{Trader} & \multirow{14}[7]{*}{News}  & You're an equity trader, and we're deliberating on the positioning for the stock of {Company} based on the latest news update. Our objective is to capitalize on potential market movements within the upcoming week. To aid our decision, it's vital to gauge whether the recent news will significantly influence investor views and cause short-term price fluctuations. Kindly evaluate the current news and determine our next steps. When providing your response, use the template below for clarity:\newline{}\newline{}[Action]: Choose either "long", "short", or "neither"  \newline{}[Thoughts]: Briefly outline your rationale.\newline{}\newline{}News Title: \newline{}{\{News Title\}}\newline{}\newline{}News Content: \newline{}{\{News Content\}} \\
    \hline
    \multirow{10}[5]{*}{Trader} & \multirow{10}[5]{*}{Analyst's Analysis} & You're an equity trader, and we're deliberating on the positioning for the stock of  {Company}. Our objective is to capitalize on potential market movements within the upcoming week. To aid our decision, it's vital to gauge whether the recent news will significantly influence investor views and cause short-term price fluctuations. \newline{}Kindly evaluate our analysts' insights and determine our next steps. When providing your response, use the template below for clarity:\newline{}[Action]: Choose either "long", "short", or "neither"\newline{}[Thoughts]: Briefly outline your rationale.\newline{}\newline{}Analyst's Analysis: \newline{}{\{Analyst's Analysis\}} \\
    \hline
    \multirow{16}[8]{*}{Trader} & \multirow{16}[8]{*}{News + Analyst's Analysis} & You're an equity trader, and we're deliberating on the positioning for the stock of {Company} based on the latest news update. \textcolor{red}{Our objective is to capitalize on potential market movements within the upcoming week. To aid our decision, it's vital to gauge whether the recent news will significantly influence investor views and cause short-term price fluctuations.}. Kindly evaluate the current news against our analysts' insights and determine our next steps. When providing your response, use the template below for clarity:\newline{}[Action]: Choose either "long", "short", or "neither"\newline{}[Thoughts]: Briefly outline your rationale.\newline{}\newline{}News Title: \newline{}{\{News Title\}}\newline{}\newline{}News Content: \newline{}{\{News Content\}}\newline{}\newline{}Analyst's Analysis: \newline{}{\{Analyst's Analysis\}} \\
    \hline
    \multirow{20}[10]{*}{Head Trader} & \multirow{20}[10]{*}{News + Analyst's Analysis + Trader's Suggestion} & As the leader of our trading desk, you carry the mantle of steering our equity trading decisions with precision and foresight. We\textbackslash{}'re deliberating on the positioning for the stock of  {Company}. Both the research department's insights and the \textcolor{red}{junior} trader's proposition have been laid out for your consideration. Your mandate is to determine the appropriateness of the \textcolor{red}{junior} trader's recommendation (Long, Short, or Neither) in the context of the prevailing market information. To communicate your decision effectively, please adhere to the template below:   \newline{}[Action]: Your options are "Follow" to proceed with the \textcolor{red}{junior} trader's suggestion, or "Not Follow" if you believe a different course is warranted. If "Follow" is your action of choice, omit the below section.\newline{}[Thoughts]: If you choose "Not Follow", briefly elucidate your reasoning.\newline{}\newline{}\textcolor{red}{Junior} trader's suggestion: \newline{}{\{Trader's suggestion\}}\newline{}\newline{}News Title: \newline{}{\{News Title\}}\newline{}\newline{}News Content: \newline{}{\{News Content\}}\newline{}\newline{}Analyst's Analysis: \newline{}{\{Analyst's Analysis\}} \\
    \end{tabular}%
    }
  \caption{Prompts for different roles.}
  \label{tab:prompts}%
\end{table*}%

\end{document}